\documentclass[10pt,twocolumn,letterpaper]{article}

\usepackage[pagenumbers]{cvpr}      

\definecolor{cvprblue}{rgb}{0.21,0.49,0.74}
\usepackage[pagebackref,breaklinks,colorlinks,allcolors=cvprblue]{hyperref}

\usepackage{cuted}
\usepackage{capt-of}
\usepackage{multirow}
\usepackage{colortbl}

\newcommand{\MYhref}[3][blue]{\href{#2}{\color{#1}{#3}}}

\def\paperID{08978} 
\def\confName{CVPR}
\def\confYear{2025}

\title{\vspace{-0.8cm} DreamColour: Controllable Video Colour Editing without Training \vspace{-0.7cm}}

\author{
\MYhref[cvprblue]{https://chaitron.github.io}{Chaitat Utintu}\textsuperscript{\thanks{Interned with SketchX}} \hspace{.2cm}
\MYhref[cvprblue]{https://www.pinakinathc.me}{Pinaki Nath Chowdhury}\textsuperscript{1} \hspace{.2cm}
\MYhref[cvprblue]{https://aneeshan95.github.io}{Aneeshan Sain}\textsuperscript{1} \hspace{.2cm}
\MYhref[cvprblue]{https://subhadeepkoley.github.io}{Subhadeep Koley}\textsuperscript{1} \\ \MYhref[cvprblue]{https://ayankumarbhunia.github.io}{Ayan Kumar Bhunia}\textsuperscript{1} \hspace{.2cm}  \MYhref[cvprblue]{https://www.surrey.ac.uk/people/yi-zhe-song}{Yi-Zhe Song}\textsuperscript{1} \\
\textsuperscript{1}SketchX, CVSSP, University of Surrey, United Kingdom.  \\
{\tt\small utintu.c@gmail.com}\\
{\tt\small \{p.chowdhury, a.sain, s.koley, a.bhunia, y.song\}@surrey.ac.uk}\\
\small\url{https://chaitron.github.io/DreamColour-demo}
}

\begin{document}
\maketitle
\begin{strip}\centering
\vspace{-54pt}
\captionsetup{type=figure}
\includegraphics[width=\textwidth]{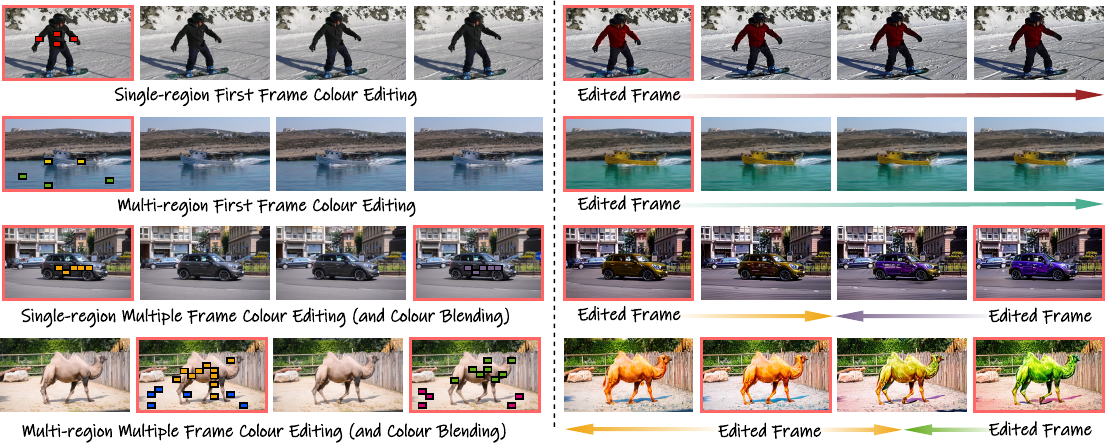}
\vspace{-6mm}
\captionof{figure}{Our training-free framework enables intuitive video colour editing in two stages. First, users simply select colours from a $16\times16$ grid to edit any frame, with automatic instance segmentation~\cite{ravi2024sam2} preventing colour bleeding. Then, our bidirectional propagation mechanism, combining temporal attention~\cite{ho2022videodiffusionmodels} and motion-aware blending~\cite{szeliski2021CV}, ensures smooth colour transitions across frames. This approach enables flexible editing scenarios: from single to multiple regions, and from any frame in the sequence, while maintaining temporal consistency through careful integration of diffusion inversion~\cite{song2022denoisingdiff} and instance-aware colour control~\cite{cong2024colourisation}.}
\vspace{-2mm}
\label{fig:feature-graphic}
\end{strip}

\begin{abstract}

\vspace{-3mm}
Video colour editing is a crucial task for content creation, yet existing solutions either require painstaking frame-by-frame manipulation or produce unrealistic results with temporal artefacts. We present a practical, training-free framework that makes precise video colour editing accessible through an intuitive interface while maintaining professional-quality output. Our key insight is that by decoupling spatial and temporal aspects of colour editing, we can better align with users' natural workflow -- allowing them to focus on precise colour selection in key frames before automatically propagating changes across time. We achieve this through a novel technical framework that combines: (i) a simple point-and-click interface merging grid-based colour selection with automatic instance segmentation for precise spatial control, (ii) bidirectional colour propagation that leverages inherent video motion patterns, and (iii) motion-aware blending that ensures smooth transitions even with complex object movements. Through extensive evaluation on diverse scenarios, we demonstrate that our approach matches or exceeds state-of-the-art methods while eliminating the need for training or specialized hardware, making professional-quality video colour editing accessible to everyone.

\end{abstract}

\vspace{-3mm}
\section{Introduction}
\label{sec:intro}
Imagine being able to change the colour of any object in a video with just a few clicks -- a red dress becoming blue across an entire fashion show, autumn leaves transforming to spring green throughout a scene, or a car changing colour smoothly as it drives past (\cref{fig:feature-graphic}). While this capability would revolutionise content creation across film, advertising, and social media, current video editing tools make such changes complex, requiring frame-by-frame edits or producing unrealistic results with temporal inconsistencies.


The fundamental challenge lies in the complexity of video colour editing: changes must be spatially precise within each frame while maintaining temporal consistency across the video, all while preserving the original lighting, textures, and motion. Current solutions force an impossible choice: traditional tools require painstaking frame-by-frame editing~\cite{truong2016QuickCut, ranjan2008television}, learning-based methods~\cite{iizuka2019deepremaster,teitelman2021transformer} produce visible artefacts and temporal flickering despite extensive training requirements, and recent automated approaches sacrifice precise control~\cite{chen2018language, wang2021Story, chang2023LCAD} or temporal consistency~\cite{yang2024BiSTNet, zhao2023SVCNet} in pursuit of usability. Even SOTA methods that achieve better results require prohibitive computational resources and days of training on paired data~\cite{ho2022videodiffusionmodels, ho2022imagenvid}, putting them out of reach for most users.

Our key insight addresses these challenges by fundamentally reframing video colour editing: instead of requiring users to simultaneously manage frame-by-frame edits and temporal consistency -- a task that has proven nearly impossible to automate effectively without extensive training \cite{vondrick2018tracking} -- we recognise that spatial and temporal aspects are inherently separable \cite{ku2024AnyV2V}. This insight enables a training-free approach that makes precise editing accessible: users can focus on intuitive colour selection in key frames, then have those edits propagate naturally across time. This separation naturally aligns with how users think about video editing \cite{lee2022PopStage, choi2016PlotAnalysis, huang2021where} while eliminating the computational overhead and data requirements that made prior methods impractical.

This insight leads to our novel technical framework that systematically addresses the video colour editing challenge. At its core, our approach begins with spatial precision through a hybrid interface combining grid-based colour selection with automatic instance segmentation. Users select colours from a simple grid interface, while our system automatically identifies and respects object boundaries -- preventing the colour bleeding artefacts that plague existing solutions. This careful handling of spatial editing provides the foundation for consistent colour propagation across frames.

Building on this spatial precision, we leverage the rich generative priors of pre-trained diffusion models for temporal coherence through bidirectional propagation, all without requiring any training or fine-tuning. Unlike previous approaches that force unidirectional colour flow~\cite{dai2024paintBucket,huang2024LVCD} or require extensive model adaptation \cite{bozic2024versatile}, our method exploits the natural motion patterns captured in pre-trained diffusion model latent space, working in both forward and backward directions. By extracting and utilising these inherent motion cues through careful attention control and latent space manipulation, we enable colour edits to propagate smoothly in both directions -- allowing edits from any frame while maintaining consistent object appearance throughout the video. 

Our motion-aware blending mechanism serves as the bridge between spatial and temporal aspects, operating directly in diffusion model feature space to ensure coherent propagation. By carefully manipulating cross-attention maps and leveraging self-attention features, our system dynamically adjusts colour propagation based on scene dynamics. When objects move quickly, the system adapts its blending strategy to maintain sharp boundaries; when motion is subtle, it smoothly interpolates colours. This adaptive behaviour in feature space ensures consistent and visually pleasing results across the entire video, regardless of motion complexity or scene changes.

Specifically, our contributions include:
\textit{(i)} A training-free framework for video colour editing that achieves professional-quality results without specialised computing resources.
\textit{(ii)} An intuitive grid-based interface with automatic instance segmentation that enables precise spatial control without frame-by-frame editing.
\textit{(iii)} A bidirectional colour propagation technique that maintains temporal consistency while allowing edits from any frame.
\textit{(iv)} A dynamic blending mechanism that ensures smooth colour transitions across complex object motions and occlusions.

\section{Related Works}
\label{sec:related}
\subsection{Diffusion-based Image and Video Generation}
Recent advancements in diffusion models (DMs) have led to state-of-the-art performance in image and video synthesis by iteratively denoising inputs to match target data distributions, showing strong generative capacity across complex domains \cite{ho2020denoisingdiff, song2021scorebasedgen, nichol2021improveddenoising, song2022denoisingdiff, dhariwal2021diffusionbeatsGAN, nichol2022glide, rombach2022ldm}. Unlike earlier VAE \cite{kingma2022vae} and GAN \cite{goodfellow2014gans} approaches, DMs benefit from stable training on large datasets, in text-to-image applications \cite{ramesh2021zeroshot,rombach2022ldm,nichol2022glide,saharia2022photorealistic}. These text prompt conditioning methods allow precise, text-driven control over generated images, enhancing flexibility and enabling guided synthesis. Extending this framework to video synthesis, DMs incorporate spatio-temporal modules that ensure temporal consistency across frames, addressing the unique challenge of maintaining motion continuity \cite{ho2022videodiffusionmodels,ho2022imagenvid,singer2022makeavideo,blattmann2023alignlatent}. Despite these advancements, video generation remains computationally intensive, requiring large annotated datasets and substantial resources, which limits rapid progress in this field. 

\subsection{Image Editing with Diffusion Models}
Several studies have extended diffusion models beyond text-conditioned image generation by incorporating additional conditioning signals for controllable image generation and image-to-image editing \cite{saharia2022palette, wang2022pretrainingneed, mou2023t2iadapter, voynov2022sketchguided, zhang2023controlnet}. Palette \cite{saharia2022palette} has demonstrated applications like colourisation, inpainting, and uncropping within a diffusion model framework. Other approaches add control signals, \textit{e.g.}, sketches, segmentation maps, or depth maps, by adapting pre-trained image generation models through methods like fine-tuning \cite{wang2022pretrainingneed}, adapter layers \cite{mou2023t2iadapter}, or trainable modules \cite{voynov2022sketchguided, zhang2023controlnet}. ControlNet \cite{zhang2023controlnet} has effectively enabled high-quality image generation from various conditions, including edge maps, depth maps, and keypoints, by fine-tuning an attached trainable copy of the diffusion model with zero-initialised convolution layers, preserving the integrity of the original model. Other methods have aimed to edit images while retaining their semantic structure through techniques such as attention layer manipulation \cite{hertz2022prompttoprompt, tumanyan2022plugandplay}, optimisation-based guidance \cite{choi2021ilvr, parmar2023zeroshotimagetoimage, epstein2023diffusionselfguidance}, or per-instance fine-tuning \cite{kawar2023imagic}. Plug-and-Play \cite{tumanyan2022plugandplay} maintains structure by integrating self-attention maps and internal features from the original image during feature reconstruction. Self-Guidance \cite{epstein2023diffusionselfguidance}, and Pix2Pix-Zero \cite{parmar2023zeroshotimagetoimage} employ a guidance loss during generation to achieve intended edits. Prompt-to-Prompt \cite{hertz2022prompttoprompt} facilitates image modification by reweighting cross-attention maps tied to different prompts. In this paper, we extend key concepts from image editing techniques to video colour editing by applying \textit{attention layer manipulation} across frames and propagating colour modifications from the initial frame while using \textit{spatio-temporal injection} to preserve the original semantic structure of a video and ensure smooth motion continuity.

\subsection{Video Editing with Diffusion Models}
Training large-scale video editing models is challenging due to scarce paired video data and high computational costs. Recent text-to-image (T2I) diffusion models \cite{rombach2022ldm} have advanced text-driven image editing \cite{avrahami2022blendeddiffusion, hertz2022prompttoprompt, tumanyan2022plugandplay, parmar2023zeroshotimagetoimage}, motivating efforts to adapt these pre-trained T2I models for video editing. However, unlike image editing, video editing must adjust appearance-based attributes while strictly preserving temporal coherence across frames. Lack of temporal consistency results in artefacts, such as flickering and frame degradation, reducing video quality and stability.

Pre-trained T2I models for video editing can be broadly classified into two approaches: i) per-video fine-tuning \cite{liu2023videop2p, shin2023editavideo, wu2023tuneavideo} and ii) zero-shot methods \cite{ceylan2023pix2video, qi2023fatezero, geyer2023tokenflow, khachatryan2023text2videozero, yang2023rerendervideo, zhang2023controlvideo}. Fine-tuning methods optimise the T2I model’s parameters for each source video, enhancing temporal coherence in the target video. For example, Tune-A-Video \cite{wu2023tuneavideo} and VideoP2P \cite{liu2023videop2p} fine-tune text-to-image models to achieve smooth motion, though these methods are computationally intensive. To address this, zero-shot methods improve temporal consistency without training by transitioning from spatial self-attention in T2I diffusion models to temporal-aware cross-frame attention with early latent fusion. For example, Pix2Video \cite{ceylan2023pix2video} and Fate-Zero \cite{qi2023fatezero} retain structural and motion details by leveraging inverted latents from text-to-image models. However, zero-shot methods often experience flickering issues due to \textit{limited temporal knowledge}.

To maintain zero-shot simplicity while addressing flickering artefacts, we propose a training-free video colour editing method that leverages the rich motion priors inherent in a pre-trained image-to-video (I2V) diffusion model \cite{zhang2023i2vgenxl}. Additionally, by utilising colour hints rather than text prompts, our approach enables precise control over colours and their locations, overcoming the limitations of natural language ambiguity and text-to-image (T2I) models \cite{rombach2022ldm}.

\section{Proposed Method}
\label{sec:method}
\paragraph{Overview.}
Our video colour editing method uses a two-stage approach that combines interactive editing with seamless colour propagation across frames. In the initial editing phase, users provide ``colour hints'' on a $16 \times 16$ grid, specifying colours and regions with precision, reducing ambiguity compared to textual prompts. Object masks generated by SAM2 \cite{ravi2024sam2} prevent colour spillover, while the Hybrid-Transformer from UniColor \cite{huang2022unicolor} ensures sharp boundaries in single-colour regions. For multi-region edits, a dual-prompt technique applies user-defined colour hints as positive prompts, with surrounding colours as negative prompts, enhancing mask and colour accuracy across selected areas.

The second stage focuses on propagating colour across frames, ensuring both spatial consistency and temporal coherence. BLIP2 \cite{li2023blip2} generates descriptive text for object colours, guiding consistent colour application. Using an I2V model (I2VGen-XL \cite{zhang2023i2vgenxl}) and DDIM inversion \cite{song2022denoisingdiff}, the edited frame is conditioned with object descriptions to synchronise colour across frames. For intermediate frames, two DDIM inversions (forward and reverse) support continuity, while a linear blend operator \cite{szeliski2021CV} computes weighted sums based on proximity to the edited frame, enabling smooth transitions. This approach provides precise, temporally consistent video colour editing without retraining.

\subsection{Preliminary}
\paragraph{Diffusion Models.}
Diffusion Probabilistic Models (DPMs) \cite{ho2020denoisingdiff,song2022denoisingdiff,dhariwal2021diffusionbeatsGAN} approximate a data distribution \(p(x)\) by progressively denoising a normally distributed variable. The denoising function learns to reverse a fixed Markov Chain process of length \(T\). Diffusion modelling involves two stochastic phases: \(forward\) and \(backward\) diffusion \cite{song2022denoisingdiff}. In training, the \(forward\) phase gradually adds Gaussian noise to an original image \(x_{0} \in \mathbb{R}^{H \times W \times 3}\), producing a noisy image \(x_{t} \in \mathbb{R}^{H \times W \times 3}\). This can be expressed as \( x_{t} = \sqrt{\bar{\alpha}_{t}} x_{0} + \sqrt{1 - \bar{\alpha}_{t}} \, \epsilon \), where \(\epsilon \sim \mathcal{N}(0, \textbf{I})\) is the Gaussian noise added, \(\alpha_{t}\) controls noise level, varying from \(\alpha_{0} = 1\) to approximately \(\alpha_{T} \approx 0\), and \(t\) is sampled uniformly from \(\{1, \ldots, T\}\) \cite{song2022denoisingdiff}. For \(backward\) diffusion, denoising autoencoders \(\epsilon_{\theta}(\cdot)\) are trained to produce a noise-free image by minimising the objective:

\vspace{-2mm}
{\small
\begin{equation}
    \mathcal{L}_\textrm{DM} = \mathbb{E}_{x, \epsilon \sim \mathcal{N}(0,1), t} \left[ \| \epsilon - \epsilon_{\theta}(x_{t}, t) \|_{2}^{2} \right]
    \vspace{-1mm}
\end{equation}}
In inference, the trained denoising autoencoder \(\epsilon_{\theta}(\cdot)\) refines a Gaussian sample \(x_{T}\) across \(T\) steps, producing a denoised image \(x_{0}\) that approximates the target data distribution \cite{song2022denoisingdiff}.

\paragraph{Latent Diffusion Models.}
Latent Diffusion Models (LDMs), \textit{i.e.,} Stable Diffusion \cite{rombach2022ldm}, shift from modelling data in high-dimensional pixel space to a more efficient low-dimensional latent space. This is achieved using an autoencoder, consisting of an encoder \(\mathcal{E}(\cdot)\) and decoder \(\mathcal{D}(\cdot)\) (typically based on a UNet backbone \cite{ronneberger2015unet}), for forward and backward diffusion \cite{rombach2022ldm}. For an input image \(x_{0}\in \mathbb{R}^{H\times W\times3}\), the encoder compresses it to a latent representation \(z_{0}\in \mathbb{R}^{h\times w\times d}\) by a factor \(f=H/h=W/w\), where \(z_{0}=\mathcal{E}(x_{0})\) \cite{rombach2022ldm}, and the decoder reconstructs it as \(\tilde{x_{0}}=\mathcal{D}(z_{0})=\mathcal{D}(\mathcal{E}(x_{0}))\). Conditional generation, \textit{e.g.,} textual prompts, is achieved by incorporating cross-attention mechanism within the UNet backbone \cite{vaswani2023attentionneed} to enable the conditional denoising network \(\epsilon _{\theta }(z_{t},t,c)\) learn \(p(z|c)\), with \(c\) as the condition embedding. The modified objective is:
\vspace{-1mm}
{\small
\begin{equation}
\label{eq2}
    \mathcal{L}_\textrm{LDM}=\mathbb{E}_{\mathcal{E}(x),c,\epsilon \sim \mathcal{N}(0,1),t}\left [ \left \| \epsilon -\epsilon _{\theta }(z_{t},t,c) \right \| _{2}^{2}\right ]
    \vspace{-2mm}
\end{equation}
}
This objective minimises latent-space reconstruction error, enabling a highly efficient conditional image generation.

\paragraph{Video Latent Diffusion Models.} Video Latent Diffusion Models (VLDMs), \textit{i.e.,} I2VGen-XL \cite{zhang2023i2vgenxl}, builds on Latent Diffusion Models (LDMs) \cite{rombach2022ldm} by introducing spatial and temporal self-attention layers into the denoising model, \(\epsilon_{\theta}(\cdot)\), typically a UNet \cite{ronneberger2015unet}. By adapting 2D convolutions to 3D, these layers enable VLDMs to capture temporal continuity, which is essential for video data. In Image-to-Video generation, given a reference frame $c_{i}$ and a guiding textual prompt $c_{t}$, the model aims to generate a video sequence \( X_0 = \{ x_0^i\}_{i=1}^N \) with smooth motion, maintaining consistency with \( c_{i} \) and \( c_{t} \). A noisy latent \( z_{t} \in \mathbb{R}^{F \times H \times W \times C} \) at each time step \(t\) (where \(F\), \(H\), \(W\), and \(C\) denote frame, height, width, and channel) is progressively denoised by \(\epsilon_{\theta}(z_{t}, t, c_{i}, c_{t})\), with the following loss function:

\vspace{-4mm}
{\small
\begin{equation}
\label{eq3}
    \mathcal{L}_\textrm{VLDM} = \mathbb{E}_{\mathcal{E}(x), c_{t}, c_{i}, \epsilon \sim \mathcal{N}(0,1), t} \left[ \left\| \epsilon - \epsilon_{\theta}(z_{t}, t, c_{t}, c_{i}) \right\|_2^2 \right]
    \vspace{-2mm}
\end{equation}
}
The noisy latent \(z_{t}\) is derived from the true latent \(z_{0}\) as \( z_{t} = {\alpha}_{t}x_{0} + {\sigma}_{t}\epsilon \), where \({\sigma}_{t} = \sqrt{1 - {{\alpha}_{t}}^2}\). Hyperparameters \({\alpha}_{t}\) and \({\sigma}_{t}\) regulate noise levels in the diffusion process, guiding the model to produce temporally consistent video frames aligned with the input conditions \(c_{i}\) and \(c_{t}\).

\subsection{Intra-Frame Colour Editing Stage}

Unlike traditional text-guided diffusion-based image editing methods \cite{chang2023LColns}, our approach introduces an interactive $16 \times 16$ grid $\mathcal{C}_{u} \in \mathbb{R}^{16 \times 16 \times 3}$ on a single frame $\mathcal{I}$ of a video $\mathcal{V} = \{ \mathcal{I}_{i}\}_{i=1}^N$. This grid enables users to ``click'' and specify precise RGB colours along with $(x_1 y_1, x_2 y_2, \dots )$ hint points, offering clear control over both colour and location. We leverage these user-provided colour hints, along with UniColor \cite{huang2022unicolor} and SAM2 \cite{ravi2024sam2}, to enable a zero-shot, user-guided image colour editing approach.


In video colour editing, precise frame colour editing is crucial, as it establishes the foundation for subsequent frame propagation. With user-defined colour hints $\mathcal{C}_{u}$ and its corresponding mask $\mathcal{M}_{\mathcal{C}_{u}} \in \mathbb{R}^{16 \times 16 \times 1}$, our method aims to achieve two core objectives for colour consistency: \textit{(i)} accurately reproducing user-specified colours in the designated regions (\ie, $colour(\mathcal{I}_{edited}\odot\mathcal{M}_{\mathcal{C}_u})\approx{C}_u$), where $\odot$ represents the Hadamard product, and \textit{(ii)} preserve colour of the non-selected areas of the frame in their original form: $colour(\mathcal{I}_{edited}\odot(1-\mathcal{M}_{\mathcal{C}_u}))\approx colour(\mathcal{I}\odot(1-\mathcal{M}_{\mathcal{C}_u})$.


\subsubsection{Single-region Frame Colour Editing}
\label{sec:single-region-editing}

\vspace{-3mm}
\begin{figure}[!htbp]
    \centering
        \includegraphics[width=\linewidth]{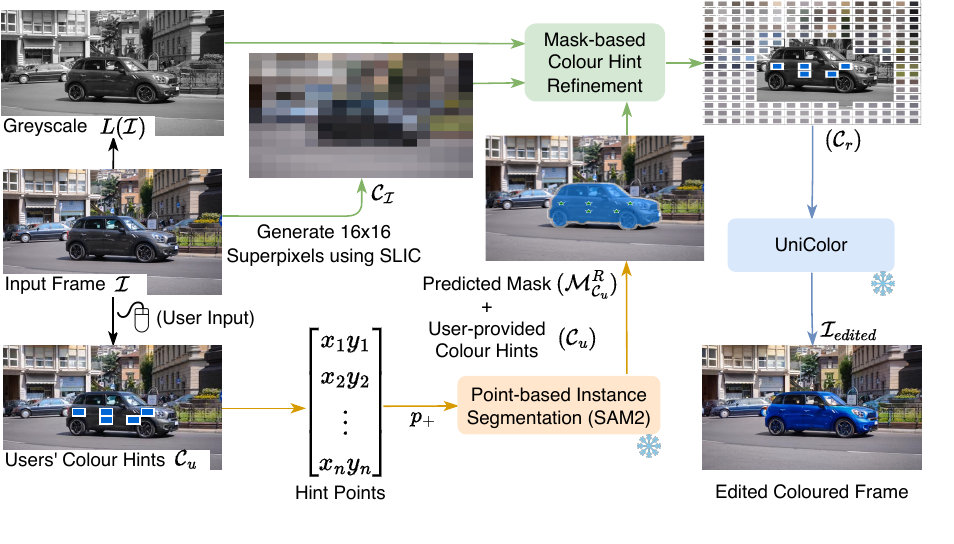}
        \vspace{-10mm}
    \caption{Our single-region colour editing begins with greyscale conversion and superpixel generation to create structural foundation and initial colour hints, respectively. User-defined hints are refined with SAM2 instance segmentation, creating an accurate object mask that guides UniColor to produce an edited frame with targeted colour applications and preserve the unselected regions.}
    \label{fig:single_colour}
    \vspace{-2mm}
\end{figure}

To achieve precise single-region colour editing, as illustrated in \cref{fig:single_colour}, we begin by transforming the input RGB frame $\mathcal{I} \in \mathbb{R}^{H \times W \times 3}$ into the CIE Lab colour space, isolating the luminance channel $L(\mathcal{I}) \in \mathbb{R}^{H \times W \times 1}$ to produce a greyscale image that serves as the structural foundation for UniColor-based colourisation. Next, we apply Simple Linear Iterative Clustering (SLIC) \cite{SLIC}, an adaptation of $k$-means clustering that efficiently generates ``superpixels'', grouping RGB pixels into perceptually coherent atomic regions. These superpixels act as colour hints $\mathcal{C}_{\mathcal{I}} \in \mathbb{R}^{{16} \times {16} \times 3}$, capturing the original colour consistency before any user edits. The combined colour hints -- user-provided $\mathcal{C}_{u}$ and original image $\mathcal{C}_{\mathcal{I}}$ -- along with luminance $L(\mathcal{I})$ could theoretically be directly input to UniColor. However, our experiments revealed that this approach caused unwanted colour spillover from $\mathcal{C}_{\mathcal{I}}$ into the user-provided regions $\mathcal{C}_{u}$, resulting in inaccurate colour placement.


To address this, each point in $\mathcal{C}_u$ is used as a point prompt for SAM2 instance segmentation, generating an object mask $\mathcal{M}_{\mathcal{C}_u}^{R}$ that outlines the selected region. Within this mask, only the colours specified by the user are retained, while hint points within a $20$-pixel Euclidean distance from the mask boundary are excluded to prevent colour leakage into adjacent areas. The refined hints $\mathcal{C}_{r}$, along with the greyscale luminance $L(\mathcal{I}_1)$, are then input to UniColor. This approach meets our objectives for targeted and precise single-region colour editing with colour consistency: \textit{(i)} producing the edited image $\mathcal{I}_{edited}$ that accurately incorporates user-defined colours within the masked area, and \textit{(ii)} preserving the original appearance of unselected regions.

\subsubsection{Multi-region Frame Colour Editing}

When users select multiple colours close to each other, there is a risk of unintended colour spillover into adjacent regions, especially if these regions are part of the same object. To address this, we extend our single-region colour editing method to handle multiple target regions simultaneously, ensuring each region is edited precisely without spillover. For each target region, represented by $(\mathcal{C}_{u}^{R1}, \mathcal{C}_{u}^{R2})$, we use both positive and negative prompts in SAM2's point-based instance segmentation. The user-specified colour hints $\mathcal{C}_{u}^{R1}$ for region $\mathrm{R1}$ act as positive prompt $p^{+}$, while surrounding colours $\mathcal{C}_{u}^{R2}$ in $\mathrm{R2}$ serve as negative prompt $p^{-}$.

\begin{figure}[!htbp]
    \centering
        \includegraphics[width=0.8\linewidth]{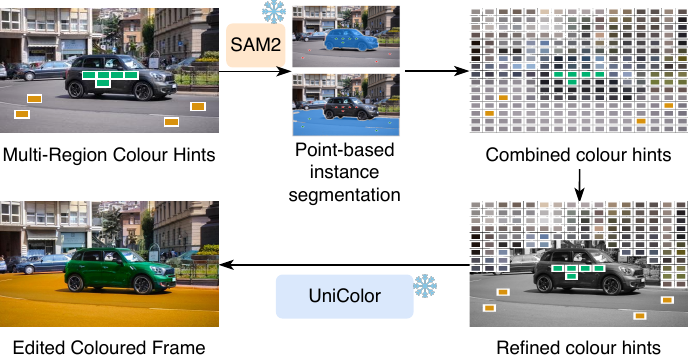}
        \vspace{-3mm}
    \caption{Multi-region colour editing pipeline using SAM2 and UniColor. SAM2's point-based instance segmentation applies positive and negative prompts to generate masks for each selected region, preventing unintended colour spillover. The combined and refined colour hints are then processed by UniColor to produce an edited frame with well-defined local colour consistency.}
    \label{fig:multi_colour}
    \vspace{-4mm}
\end{figure}

As illustrated in \cref{fig:multi_colour}: \textit{(i)} the five green user-specified colours hints are used as positive prompts $p^{+}$, and the four orange hints as negative prompts $p^{-}$ to generate the mask $\mathcal{M}_{\mathcal{C}_{u}}^{R1}$; \textit{(ii)} the four orange hints then act as positive prompts $p^{+}$, and the five green hints as negative prompts $p^{-}$ to create the mask $\mathcal{M}_{\mathcal{C}_{u}}^{R2}$. Finally, these refined masks $\mathcal{M}_{\mathcal{C}_{u}}^{R1}$ and $\mathcal{M}_{\mathcal{C}_{u}}^{R2}$ give the updated colour hints $\mathcal{C}_{r}$, which are input to UniColor to generate a multi-region edited colour frame.

This dual-prompt approach enhances boundary precision by clearly distinguishing neighbouring colours, preventing bleed across boundaries. While combining positive and negative prompts does not achieve perfect segmentation for extremely close colour hints, it produces a reasonably accurate segmentation in most cases, even when adjacent regions are part of the same object. This significantly reduces unintended blending in closely spaced, multi-coloured areas.

\vspace{-3mm}

\subsection{Inter-frame Colour Editing Stage}

This section outlines our inter-frame colour editing stage, where we transfer the colour-edited features from the initial frame $\mathcal{I}_{\text{edited}}$ to subsequent frames $\mathcal{I}_{2}, \mathcal{I}_{3}, \dots, \mathcal{I}_{n}$. Using the pre-trained I2VGen-XL \cite{zhang2023i2vgenxl} model with video generative priors, our approach requires no additional training and incorporates BLIP-2 \cite{li2023blip2} for enhanced scene semantics. A spatio-temporal feature injection method maintains structural and motion consistency across frames. Beyond first-frame editing, we support intermediate-frame editing, allowing any frame to serve as the reference, and multi-frame editing for harmonious colour blending. The key goals are: \textit{(i)} colour consistency with the edited first frame, \textit{(ii)} preservation of the original video's appearance and motion, and \textit{(iii)} temporal consistency to minimise flickering.

\subsubsection{First-frame Colour Editing}
\label{sec:first-frame-editing}

Our approach to first-frame colour editing uses two parallel I2V sampling pathways, designed to ensure accurate and consistent colour transfer across video frames.

In the primary pathway, we begin with the input video $\mathcal{V}$ and invert it into a latent noise representation $z_{t}^{\mathcal{V}} \in \mathbb{R}^{16 \times 4 \times H' \times W'}$ at time $t$ using DDIM inversion \cite{song2022denoisingdiff}. This inversion is conditioned on the first frame $\mathcal{I}_{1}$ enabling us to capture the video in the model's latent space. Operating in the latent space facilitates the manipulation of complex features like colour, structure, and motion more effectively \cite{park2023riemannian}. We then apply DDIM sampling to progressively denoise this latent representation, while simultaneously extracting spatio-temporal features from the I2V model's decoder layers. These features capture essential semantic details, such as structure and motion, which are crucial for preserving its original appearance and dynamics.

\begin{figure}[!htbp]
    \centering
        \includegraphics[width=\linewidth, trim={0mm 0mm 12mm 0mm}, clip]{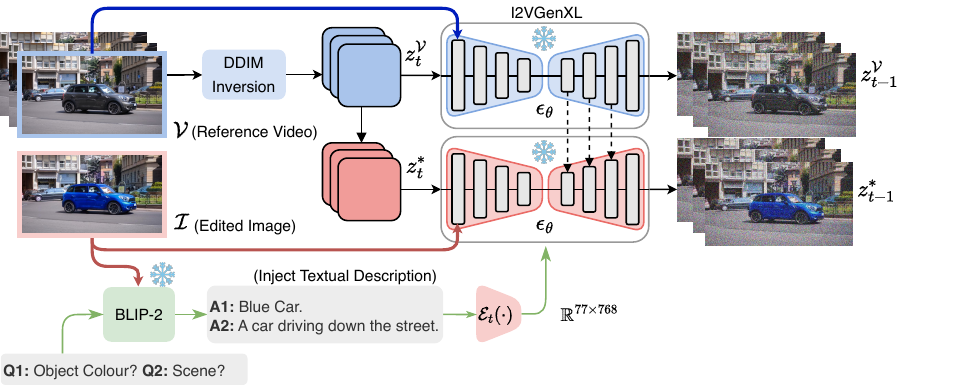}
        \vspace{-7mm}
    \caption{The primary pathway (top) performs DDIM inversion on the reference video $\mathcal{V}$, generating latent noise $z_{t}^{\mathcal{V}}$ to capture motion and structural cues. The secondary pathway (bottom) starts with the edited frame $\mathcal{I}_{edited}$ and random noise $z_{t}^{\ast}$, injecting spatio-temporal features from the primary pathway for coherence. BLIP-2 provides textual descriptions, enhancing semantic consistency and colour fidelity in the generated video.
    }
    \label{fig:first_frame}
    \vspace{-4mm}
\end{figure}

In the secondary pathway, we take the edited first frame $\mathcal{I}_{edited}$ and textual cues derived from BLIP-2 as inputs to the I2V generation model I2VGen-XL \cite{zhang2023i2vgenxl}. Starting with a random Gaussian noise $z_{t}^{\ast}$, we perform DDIM sampling while injecting the spatio-temporal features from the primary pathway (\cref{fig:first_frame}). This injection process is guided by the inverted latent representation from the original video $\mathcal{V}$, ensuring that the generated video $\mathcal{V}^{\ast}$ maintains the motion dynamics of the source video. The secondary pathway thus incorporates the structural information and motion patterns from the primary pathway, while also integrating the colour and semantic cues from the edited first frame $\mathcal{I}_{edited}$ and the additional BLIP-2 guidance from the textual prompt.

This dual-pathway approach enables us to achieve high levels of semantic fidelity and visual coherence, ensuring that the colour changes appear consistent and natural across frames, and that the edited video retains the overall appearance and motion of the source video.

\paragraph{DDIM Inversion.} To maintain frame consistency, we apply DDIM inversion to extract latent noise at each time step \( t \) from the source video \( \mathcal{V} = \{\mathcal{I}_1, \mathcal{I}_2, \dots, \mathcal{I}_n\} \), as:
\[
z_t = \texttt{DDIM\_Inv}(\epsilon_\theta(z_{t+1}, \mathcal{I}_1, \varnothing, t)),
\]
where \(\texttt{DDIM\_Inv}(\cdot)\) denotes the inversion process. The final latent noise \( z_T \) serves as the starting noise for generating edited frames, ensuring temporal coherence with the original video’s structure.

\vspace{-4mm}
\paragraph{Spatio-Temporal Feature Injection.} To maintain the appearance and motion of the source video, our approach employs spatio-temporal features \cite{tumanyan2022plugandplay, ku2024AnyV2V} comprising of convolutional, spatial attention, and temporal attention features. Spatial Feature Injection preserves background details by incorporating convolutional and spatial attention features from the denoising UNet during video sampling; specifically, DDIM-inverted latents \( z^{\mathcal{V}}_{t} \) are used to retain convolutional features \( f^{l}_{1} \) and spatial self-attention scores \( A^{l}_{2} \), parameterised by queries \( Q^{l}_{2} \) and keys \( K^{l}_{2} \), within the initial sampling stages, controlled by thresholds \( \tau_{\text{conv}} \) and \( \tau_{\text{sa}} \). Temporal Feature Injection, aimed at addressing motion consistency, integrates temporal attention features from decoder layer queries \( Q^{l}_{3} \) and keys \( K^{l}_{3} \), effectively capturing the motion of the source video within early steps, governed by \( \tau_{\text{ta}} \). By combining these spatial and temporal features, we synchronise elements in the editing branch \( \{f^{*l}_{1}, Q^{*l}_{2}, K^{*l}_{2}, Q^{*l}_{3}, K^{*l}_{3}\} \) with those of the source denoising branch, enabling a tuning-free adaptation of the I2V model for enhanced video colour editing.

\begin{figure}[b]
    \vspace{-2mm}
    \centering
        \includegraphics[width=\linewidth]{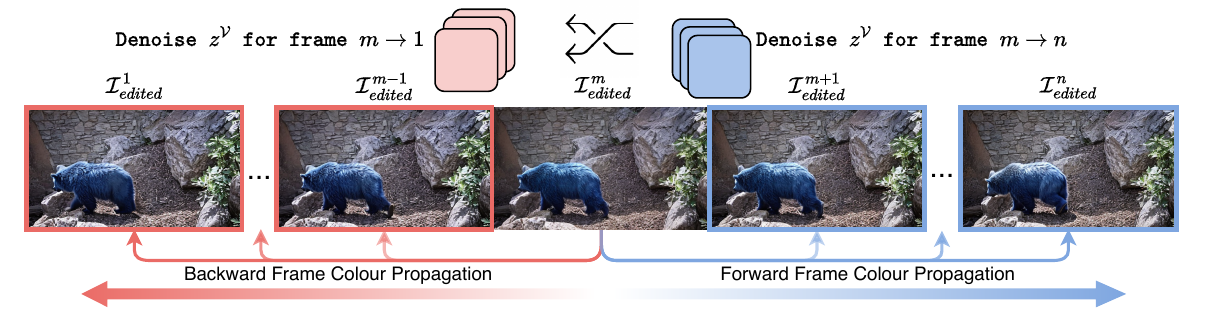}
        \vspace{-7mm}
    \caption{To edit the $m^{th}$ intermediate frame, the video is divided into forward ($\mathcal{I}_{m} \to \mathcal{I}_{n}$) and backward ($\mathcal{I}_{m} \to \mathcal{I}_{1}$) subsequences. First-frame colour editing is then applied separately in each direction, with colour changes propagated through denoising steps. The edited segments are then combined to create a fully edited video.}
    \label{fig:intermediate_frame}
\end{figure}

\vspace{-4mm}
\paragraph{VisualQA for Semantics Guidance.} To improve visual fidelity and colour consistency in edited videos, we use BLIP-2’s visual question answering \cite{li2023blip2} within an Image-to-Video (I2V) model to interpret and propagate colours across frames. By posing questions about the initial edited frame (\ie, \texttt{Please describe object colours in the scene} and \texttt{Please describe the scene}), we extract essential colour and context details, which serve as positive prompts during the DDIM sampling, as shown in \cref{fig:first_frame}. Negative prompts, such as \texttt{desaturated colour, greyish, unrealistic, ...}, counter unwanted artefacts, ensuring accurate, realistic, and aesthetically cohesive results.

\subsubsection{Intermediate Frame Colour Editing}
\label{subsec:intermediate_frame}
In the I2V generation model, the initial frame is used as the default conditional signal to propagate features across subsequent frames. However, in a video colour editing task, this constraint can limit user flexibility and creativity. Observing that reversed video sequences often present coherent, semantically inverted actions (\eg, a person standing up appears as sitting down when reversed) \cite{cachay2023DYffusion}, we introduce an intermediate frame colour editing approach. Specifically, to edit the \(m^{th}\) frame within a sequence of \(n\) frames, we segment the original video into two subsequences: \(\{{\mathcal{I}}_{m}, {\mathcal{I}}_{m+1}, \dots, {\mathcal{I}}_{n}\}\) and \(\{{\mathcal{I}}_{m}, {\mathcal{I}}_{m-1}, \dots, {\mathcal{I}}_{1}\}\). We then apply standard first-frame colour editing separately in forward and backward directions, propagating colour changes across each subsequence (see \cref{fig:intermediate_frame}), and finally, combine the edited segments to construct the output video.

\vspace{-2mm}
\subsubsection{Multiple Frame Colour Editing}
\label{sec:multiple-frame-colour-editing}
\begin{figure}[t]
    \centering
        \includegraphics[width=0.9\linewidth, trim={7mm 0mm 0mm 0mm}]{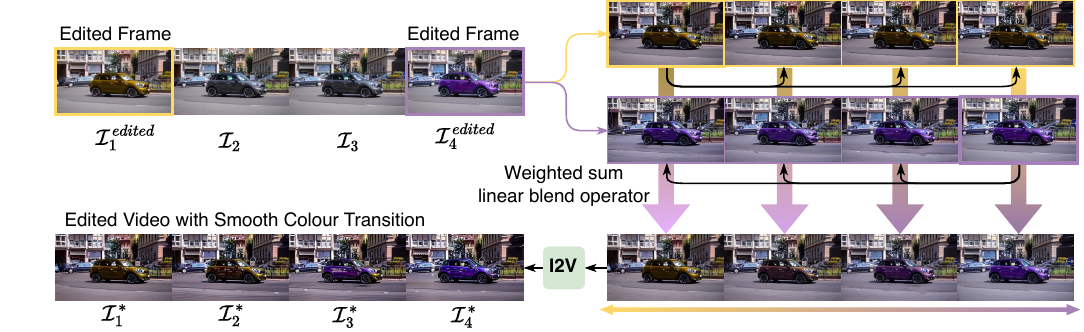}
        \vspace{-3mm}
    \caption{Smooth colour transition using intermediate frame editing. Two frames are independently edited and used for forward-backward colour propagation. A weighted sum (linear blend) merged results, followed by DDIM inversion and resampling with a guided prompt for a seamless colour blending across the video.}
    \label{fig:multi_frame}
    \vspace{-6mm}
\end{figure}

In our proposed approach (\cref{subsec:intermediate_frame}), each frame in a video can now be edited and used as a conditional signal for video colour modification. We then explore the pre-trained I2V model’s capacity to edit and seamlessly blend colours across the temporal axis. Specifically, we select two frames from the video (\eg, frame 1 and frame 4), applying distinct colour edits to each. Subsequently, we employ a forward-backward colour propagation method (see \cref{fig:intermediate_frame}) to obtain independent results from each edited frame. These frames are then merged using a weighted sum (\textit{i.e.,} a linear blend operator \cite{szeliski2021CV}), to yield a set of colour-blended frames. However, this alone does not demonstrate the blending capability of the video diffusion model. Therefore, we apply DDIM inversion to the colour-weighted frames and perform resampling with a guiding prompt, such as \texttt{a smooth colour transition across the entire scene}. The results show that the I2V model, when guided by both colour and text prompts, can effectively edit and achieve a smooth colour transition across the time axis, as depicted in \cref{fig:multi_frame}.

\vspace{-2mm}
\section{Experiments}
\label{sec:exp}
%


\subsection{State-of-the-Art Comparison}

\begin{figure}[b]
    \centering
        \includegraphics[width=\linewidth]{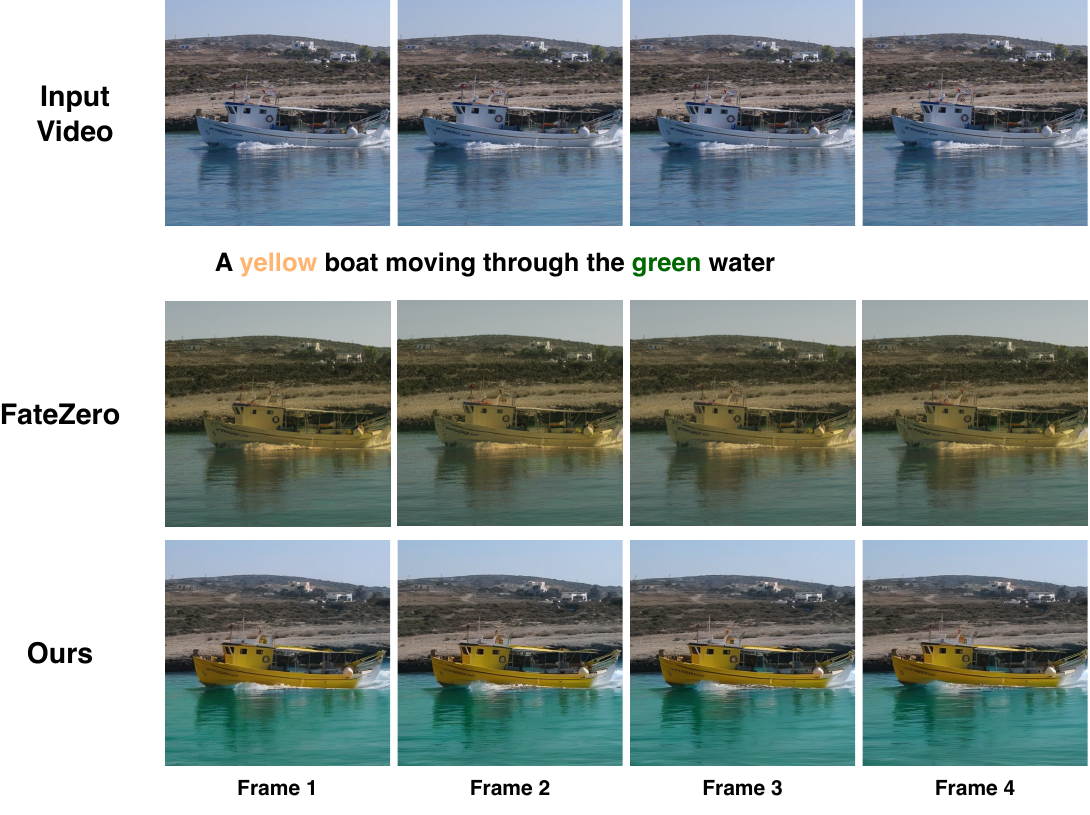}
        \vspace{-7mm}
    \caption{Comparison of our method with FateZero \cite{qi2023fatezero} for zero-shot text-based video editing. Both methods apply transformations (\eg, changing the boat to yellow and the water to green) and propagate edits across frames. FateZero, which uses textual prompts, show colour spillover and faded results. In contrast, our method with a $16 \times 16$ grid for specifying precise local colour hints maintains sharp colour boundaries and consistent local colours.}
    \label{fig:sota-compare}
\end{figure}

We compare our proposed method with FateZero \cite{qi2023fatezero}, a zero-shot text-based video editing. It combines a cross-frame attention fusion and self-attention enhancement techniques within a pre-trained diffusion model. Cross-frame attention captures semantic relationships across frames, ensuring consistent edits through the video, while self-attention helps preserve structural and motion continuity. Similar to ours, FateZero applies transformations, such as stylistic changes to one frame and propagates the edits temporally, avoiding common issues like flickering or loss of continuity across frames. However, unlike ours, which provides users an intuitive $16 \times 16$ grid to specify local colours, FateZero relies on ambiguous textual prompts. This leads to colour spillovers, as shown in \cref{fig:sota-compare}, where the yellow colour of the boat and green colour of the water, both seem to be faded. Our proposed method can preserve local colour consistency with sharp colour boundaries, thanks to the use of SAM2-based colour masks.

\subsection{Ablation Study}
\paragraph{Off-the-shelf UniColor versus SAM2-guided UniColor.}
Our proposed method uses UniColor \cite{huang2022unicolor} to generate region colour from a $16 \times 16$ colour hints. However, as mentioned in \cref{sec:single-region-editing}, using user-guided colour hints $\mathcal{C}_{u}$ as input to off-the-shelf UniColor leads to unwanted colour spillover from the surrounding region $\mathcal{C}_{\mathcal{I}}$ into the user-provided regions, resulting in inaccurate colour placement. Creating a refined colour hint $\mathcal{C}_{r}$, where each point in $\mathcal{C}_{u}$ is used as a point prompt for SAM2 instance segmentation excludes colour leakage from adjacent areas. This shows the clear benefit of our modified SAM2-guided UniColor over off-the-shelf UniColor for precise local colour edits.


\begin{figure}[t]
    \centering
        \includegraphics[width=\linewidth]{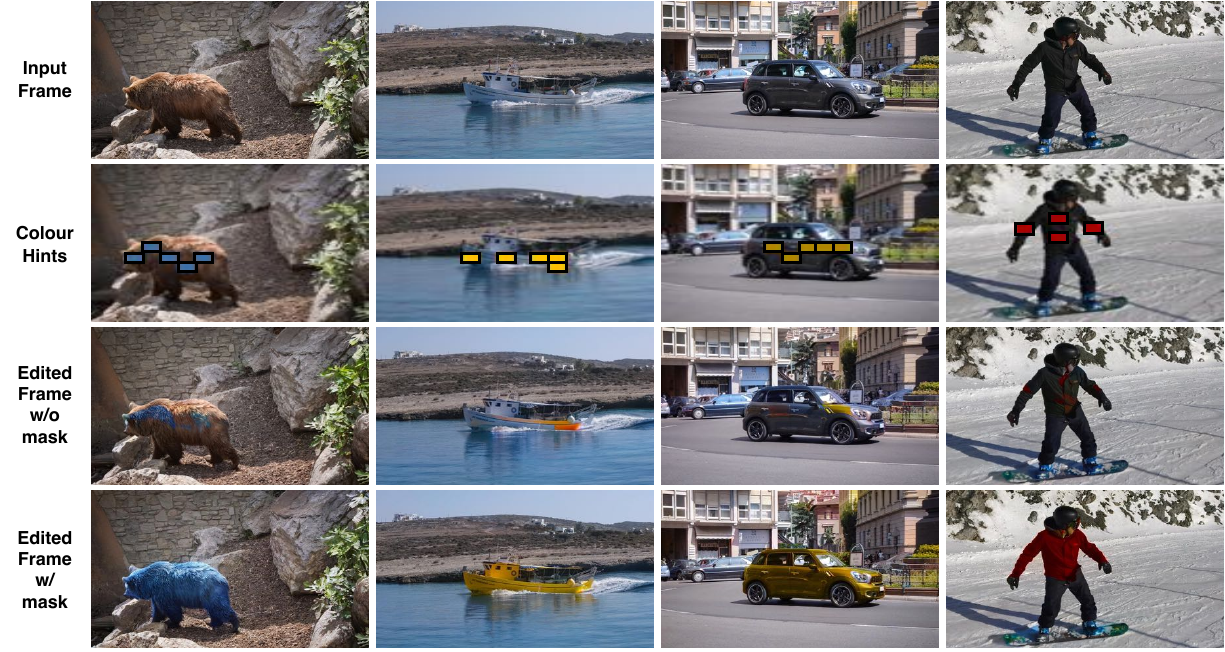}
        \vspace{-7mm}
    \caption{Ablation study on the importance of SAM2-guided UniColor over off-the-shelf for accurately reproducing user-specified colours in the designated regions.}
    \label{fig:instance-aware}
    \vspace{-5mm}
\end{figure}

\begin{figure}[ht]
    \centering
        \includegraphics[width=\linewidth]{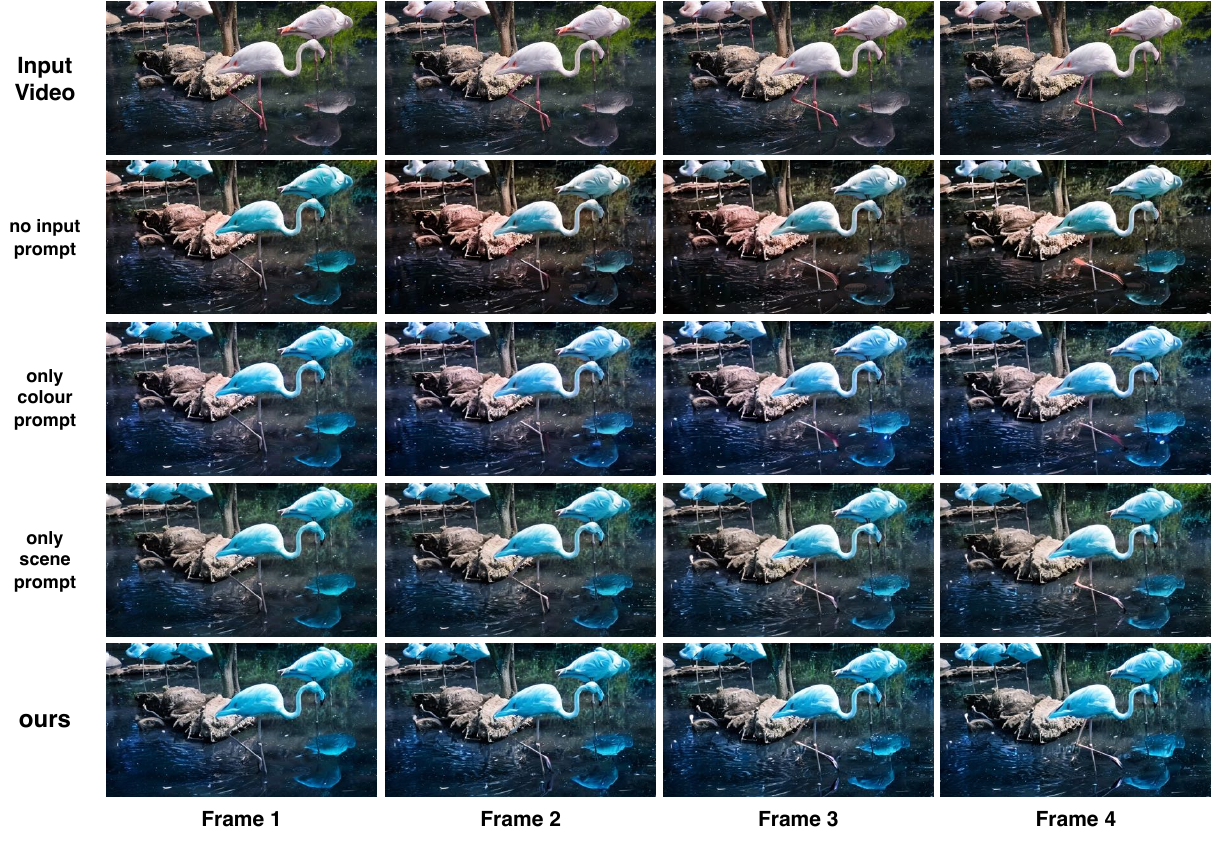}
        \vspace{-7mm}
    \caption{Ablation study on the importance of textual prompt guidance from BLIP-2 for temporal consistency and semantic fidelity.}
    \label{fig:ablation-textual}
    \vspace{-3mm}
\end{figure}

\vspace{-2mm}
\paragraph{Importance of Textual prompt guidance.}
In \cref{fig:ablation-textual}, we examine the significance of incorporating colour and scene prompts, derived from the BLIP-2 VisualQA task in \cref{sec:first-frame-editing}, during the final DDIM inversion and resampling step. \textit{(i)} First, removing all textual guidance from the BLIP-2 (second row in \cref{fig:ablation-textual}), the model struggles to comprehend the scene accurately, resulting in outputs where crucial details, such as the flamingo's legs, are absent, and the video appears temporally inconsistent. \textit{(ii)} When only the colour prompt is applied, the output exhibits enhanced saturation; however, key semantic elements, such as the flamingo's legs, remain missing (see frame 2). \textit{(iii)} Conversely, integrating only the scene prompt improves semantic fidelity, but the colour consistency degrades over the temporal axis. \textit{(iv)} Finally, injecting both colour and scene prompts, in our method, achieves coherent and visually accurate outputs, highlighting the complementary roles of these prompts.

\vspace{-2mm}
\paragraph{Can Diffusion Models Propagate Colour Backward?}
A key assumption of our intermediate frame colour editing step in \cref{subsec:intermediate_frame} is the flexibility of pre-trained diffusion models to generate videos temporally forward and backward. To assess the viability of backward colour propagation, we conducted an ablation study comparing it with the traditional forward-direction approach. In this analysis, we began with the edited first frame, compared it to the edited 15th frame, and propagated the edits backwards through the sequence. The results indicate that while backward propagation can introduce minor shifts in colour accuracy and occasional artefacts, particularly in earlier frames (\eg, frames 1 and 2), it remains a feasible and effective method. This validates the significance of backward colour propagation as a key feature of our proposed method.
\begin{figure}[b]
    \centering
        \vspace{-3mm}
        \includegraphics[width=\linewidth]{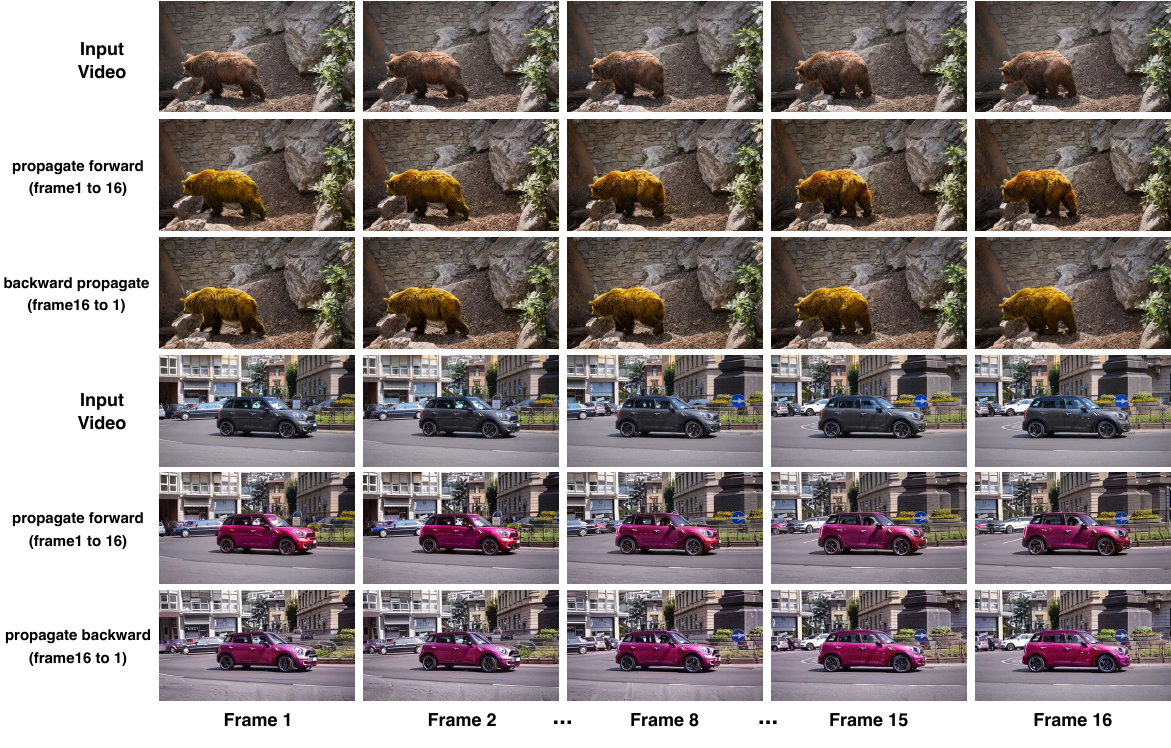}
        \vspace{-7mm}
    \caption{Ablation study on the flexibility of pre-trained video diffusion for backward colour propagation.}
    \label{fig:abla3}
\end{figure}

\vspace{-2mm}
\paragraph{Understanding Smooth Colour Transitions.}
In our final ablation study, we investigate the significance of both the weighted sum approach and the blending prompt, as detailed in \cref{sec:multiple-frame-colour-editing}, to achieve seamless multi-frame blending. The results, illustrated in the corresponding figure, underscore the necessity of both components for successful blending. For instance, in the second row, where the first two frames depict a yellow car and the last two frames transition to a purple car, the use of the blending prompt alone fails to ensure smooth integration across frames. Similarly, in the third row, the weighted sum method is applied without the blending prompt, yet the blending remains unsuccessful. These observations demonstrate that the absence of either component disrupts the blending process, highlighting the critical role of our proposed methodology in achieving harmonious transitions between multiple frames.
\begin{figure}[ht]
    \centering
        \includegraphics[width=\linewidth]{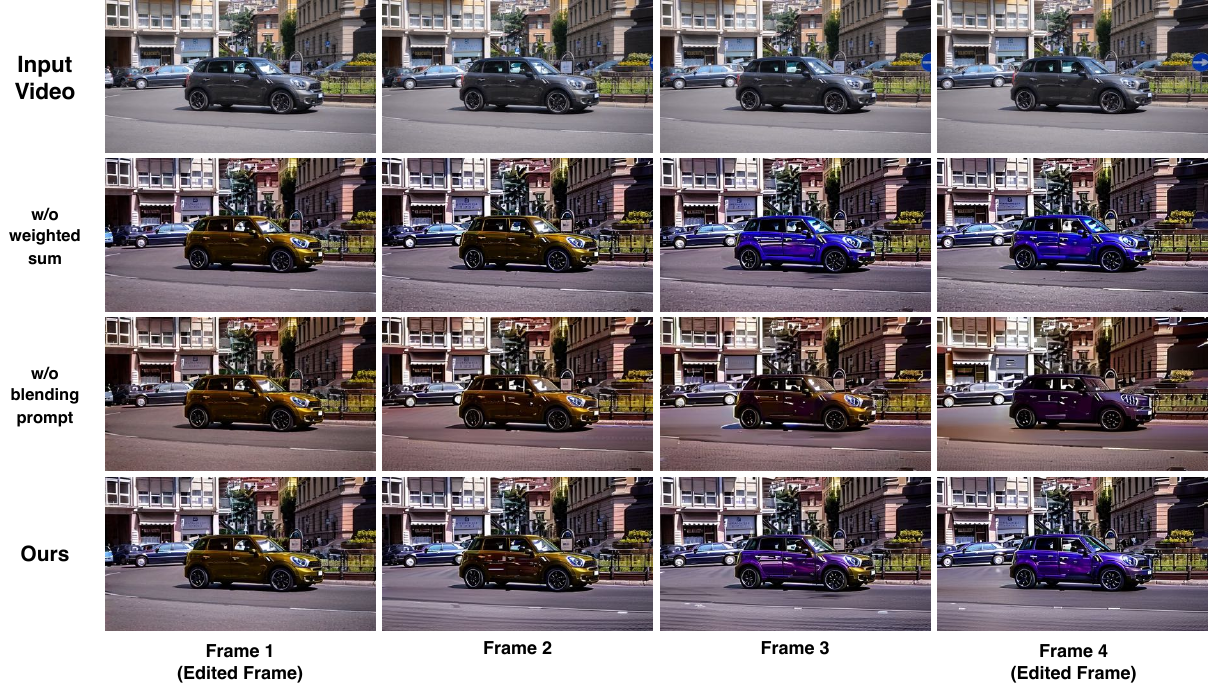}
        \vspace{-7mm}
    \caption{For smooth colour transitions across multiple edited frames, \textit{(i)} removing weighted sum and only using DDIM inversion fails to blend colours, \textit{(ii)} removing blending textual prompts disrupts the harmonious transitions.}
    \label{fig:abla4}
    \vspace{-2mm}
\end{figure}

\section{Limitations and Colour Bleeding}
\begin{figure}[ht]
    \centering
        \includegraphics[width=\linewidth]{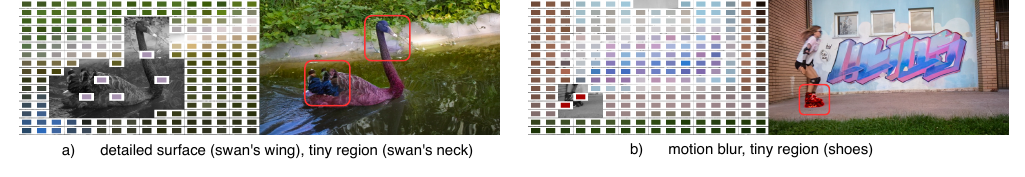}
        \vspace{-7mm}
    \caption{Limitations of our proposed method, especially of thin or fast moving objects in a video that leads to colour bleeding.}
    \label{fig:limit}
    \vspace{-2mm}
\end{figure}

The limitations of our proposed method are three-fold. First, as illustrated by the swan’s wing in Fig.\ref{fig:limit}a, our approach struggles with accurately colouring objects or regions featuring intricate textures, which resist being represented by a single colour. Second, as shown in the swan's neck in Fig.\ref{fig:limit}a and the girl's shoes in Fig.\ref{fig:limit}b, small objects or regions that are significantly smaller than the provided colour hints may result in colour bleeding or artefacts. Lastly, as demonstrated by the girl's shoes in Fig.\ref{fig:limit}b, areas affected by motion blur pose significant challenges, making it difficult to colourise such features effectively.
 
\section{Conclusion}
\label{sec:conclude}
In conclusion, DreamColour redefines video colour editing by making professional-quality results accessible without training or specialised hardware. Our key technical innovation -- decoupling spatial and temporal aspects while leveraging pre-trained diffusion models through careful attention control and dynamic blending -- enables precise colour manipulation with unprecedented temporal coherence. This training-free approach, combining instance-aware colour control with bidirectional propagation, achieves high-quality results across diverse scenarios from single-object edits to complex multi-object manipulations. Through evaluation on real-world videos, we demonstrate that our method not only matches state-of-the-art results but allows users to start editing videos intuitively through a simple point-and-click interface, without any training delays or setup time. As video content creation continues to grow across social media and entertainment, DreamColour opens new creative possibilities by bringing sophisticated colour editing capabilities to everyone.

{
    \small
    \bibliographystyle{ieeenat_fullname}
    \bibliography{main}
}

\clearpage


\twocolumn[{\centering{\Large \textbf{Supplementary material for\\ DreamColour: Controllable Video Colour Editing without Training}\par}\vspace{0.3cm}
	{\MYhref[cvprblue]{https://chaitron.github.io}{Chaitat Utintu} \hspace{.2cm}
\MYhref[cvprblue]{https://www.pinakinathc.me}{Pinaki Nath Chowdhury}\textsuperscript{1} \hspace{.2cm}
\MYhref[cvprblue]{https://aneeshan95.github.io}{Aneeshan Sain}\textsuperscript{1} \hspace{.2cm}
\MYhref[cvprblue]{https://subhadeepkoley.github.io}{Subhadeep Koley}\textsuperscript{1} \\ \MYhref[cvprblue]{https://ayankumarbhunia.github.io}{Ayan Kumar Bhunia}\textsuperscript{1} \hspace{.2cm}  \MYhref[cvprblue]{https://www.surrey.ac.uk/people/yi-zhe-song}{Yi-Zhe Song}\textsuperscript{1} \\
\textsuperscript{1}SketchX, CVSSP, University of Surrey, United Kingdom.  \\
{\tt\small utintu.c@gmail.com}\\
{\tt\small \{p.chowdhury, a.sain, s.koley, a.bhunia, y.song\}@surrey.ac.uk}\\
\small\url{https://chaitron.github.io/DreamColour-demo}\par\vspace{0.5cm}}}
]

\renewcommand\thesection{\Alph{section}}
\setcounter{section}{0}

\setcounter{figure}{0}
\renewcommand{\figurename}{Fig.}
\renewcommand{\thefigure}{S\arabic{figure}}

\setcounter{table}{0}
\renewcommand{\tablename}{Table}
\renewcommand{\thetable}{S\arabic{table}}

\section{Introduction}
\label{sec:introduct}

\noindent This supplementary material complements the main paper, "DreamColour: Controllable Video Colour Editing without Training," by providing additional experiments and details: additional quantitative and qualitative results (Sec. \ref{sec:evaluations}), ablation study on initial latent index (Sec. \ref{sec:abla2}), and clarification on contributions (Sec. \ref{sec:contribution}).

\section{Additional Performance Evaluations}
\label{sec:evaluations}

\paragraph{Dataset.} 
We evaluate our video editing performance on the DAVIS dataset \cite{ponttuset2017davis}, a benchmark widely recognised in the research community for its application in both video editing \cite{qi2023fatezero, li2023vidtome} and video colourisation \cite{liu2023videocolorization} tasks. The DAVIS dataset is particularly suited for such evaluations due to its high-quality, densely annotated video sequences that span a diverse range of scenes and motion patterns. 

\paragraph{Metrics.} 
Our task of video colour editing aims to preserve the original background colours while enhancing and evaluating the vibrancy of the edited regions, partially related to video colourisation task \cite{liu2023videocolorization}. We begin with the Fr\'echet Inception Distance (FID) \cite{heusel2018fid}, which measures perceptual realism by comparing the colour distributions of edited frames to the ground truth. LPIPS \cite{zhang2018unreasonable} evaluates perceptual similarity, offering insights into visual fidelity. Colourfulness \cite{Hasler2003MeasuringCI} assesses the colour vividness of the edited frames, aligning with human visual perception. Temporal consistency is measured using the Colour Distribution Consistency (CDC) index \cite{liu2021cdc}, which computes the Jensen-Shannon divergence of colour distributions between consecutive frames. Additionally, PSNR and SSIM are used to further analyse the structural integrity and overall perceptual quality of the edited videos.

\paragraph{Baselines.}
We structured our experiments into two distinct stages: intra-frame and inter-frame colour editing. For the intra-frame colour editing stage, we present a qualitative comparison between our method and three state-of-the-art (SOTA) image editing techniques: Plug-and-Play \cite{tumanyan2022plugandplay}, LEDITS++ \cite{brack2024ledits}, and InstructPix2Pix \cite{brooks2023instructpix2pix}. This evaluation demonstrates our method’s ability to produce high-quality colour edits on the initial frame, which serves as a critical foundation for subsequent inter-frame colour editing.

In the inter-frame colour editing stage, we provide both quantitative and qualitative comparisons against three SOTA video editing approaches: FateZero \cite{qi2023fatezero}, VidToMe \cite{li2023vidtome}, and AnyV2V \cite{ku2024AnyV2V}. The first two methods are based on text-to-image (T2I) diffusion models, while the third employs a two-stage approach based on image-to-video (I2V) diffusion which requires further adoption of first-frame editing methods, i.e., InstructPix2Pix \cite{brooks2023instructpix2pix}. The T2I-based video editing competitors can describe the robustness of our pipeline for video colour editing, whereas the I2V-based approach emphasises the importance of our intra-frame editing stage in maintaining consistency and high-quality colour transitions across frames.

\begin{figure}[hbt]
\centering
\includegraphics[width=\linewidth]{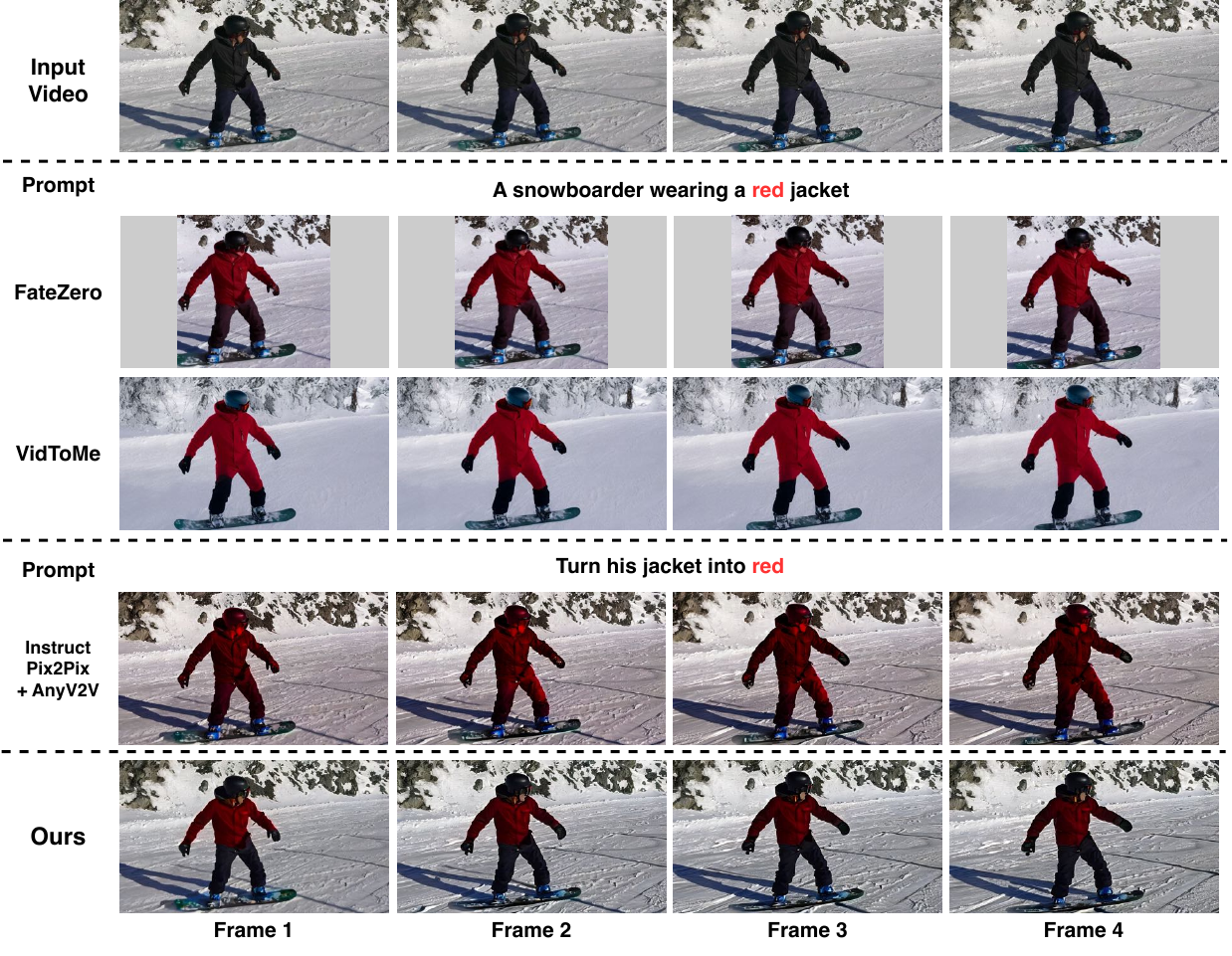}
\vspace{-6mm}
\caption{Qualitative results for inter-frame colour editing: Qualitative comparison of inter-frame colour editing stage between FateZero \cite{qi2023fatezero}, VidToMe \cite{li2023vidtome}, InstructPix2Pix \cite{brooks2023instructpix2pix} + AnyV2V \cite{ku2024AnyV2V}, and our proposed method on DAVIS \cite{ponttuset2017davis} dataset.}
\label{fig:vid}
\vspace{-4mm}
\end{figure}

\paragraph{Quantitative Evaluation.} 
In the video colour editing task, our method demonstrates superior performance compared to FateZero \cite{qi2023fatezero}, VidToMe \cite{li2023vidtome}, and AnyV2V \cite{ku2024AnyV2V}, as shown in Tab. \ref{table:quatitative}. While these baselines are primarily designed for broader video editing tasks, such as video stylisation or subject-driven editing, their adaptation to the downstream task of colour editing exposes significant limitations (see Fig. \ref{fig:vid}). Our method achieves consistently lower FID and LPIPS values, indicating enhanced visual fidelity and perceptual quality, and outperforms in SSIM and PSNR, validating the structural accuracy and pixel-level precision of our edits. Additionally, we excel in metrics such as Colorfulness and CDC, demonstrating our ability to maintain vibrant and temporally consistent colour transitions across frames, where the baselines often exhibit flickering or oversaturation during sampling.

One key objective of our framework is to preserve the semantics and integrity of unedited regions, particularly the background, ensuring temporal coherence and alignment with the edited regions. This is a significant advantage over baseline methods, which frequently introduce unwanted artefacts or distortions in unedited areas. Our higher SSIM and CDC scores reflect this ability to maintain temporal stability while ensuring that the background remains faithful to the original video. These results underscore the robustness of our framework, which leverages pre-trained modules and optimised design choices, such as weighted blending operations, to harmoniously integrate edited and unedited regions without requiring task-specific training or fine-tuning.

\begin{table}[h]
\begin{center}
\tiny
\renewcommand{\arraystretch}{1.5}
\setlength{\tabcolsep}{4pt}
\caption{Additional Quantitative Evaluation.}
\vspace{-2mm}
\scalebox{0.95}{
\begin{tabular}{ccccccc}
\hline
\rule{0pt}{3ex}
\multirow{2}{*}{\textbf{Methods}} & \multicolumn{6}{c}{\textbf{DAVIS Dataset} \cite{ponttuset2017davis}} \\ 
& \textbf{FID $\downarrow$} & \textbf{LPIPS $\downarrow$} & \textbf{Colorfulness $\uparrow$} & \textbf{CDC $\downarrow$} & \textbf{PSNR $\uparrow$} & \textbf{SSIM $\uparrow$} \\ 
\hline
FateZero \cite{qi2023fatezero} & 355.06 & 0.806 & 38.78 & 0.005278 & 8.71 & 0.280 \\
VidToMe \cite{li2023vidtome} & 351.54 & 0.791 & 39.52 & 0.005494 & 8.57 & 0.359 \\
InstructPix2Pix \cite{brooks2023instructpix2pix} + AnyV2V \cite{ku2024AnyV2V} & 395.53 & 0.762 & 25.62 & 0.003515 & 9.05 & 0.179 \\
\rowcolor{YellowGreen!40} \textbf{Proposed} & 143.83 & 0.385 & 40.53 & 0.002770 & 14.46 & 0.731 \\
\hline
\end{tabular}
}
\label{table:quatitative}
\end{center}
\end{table}

\begin{figure}[hbt]
\centering
\includegraphics[width=\linewidth]{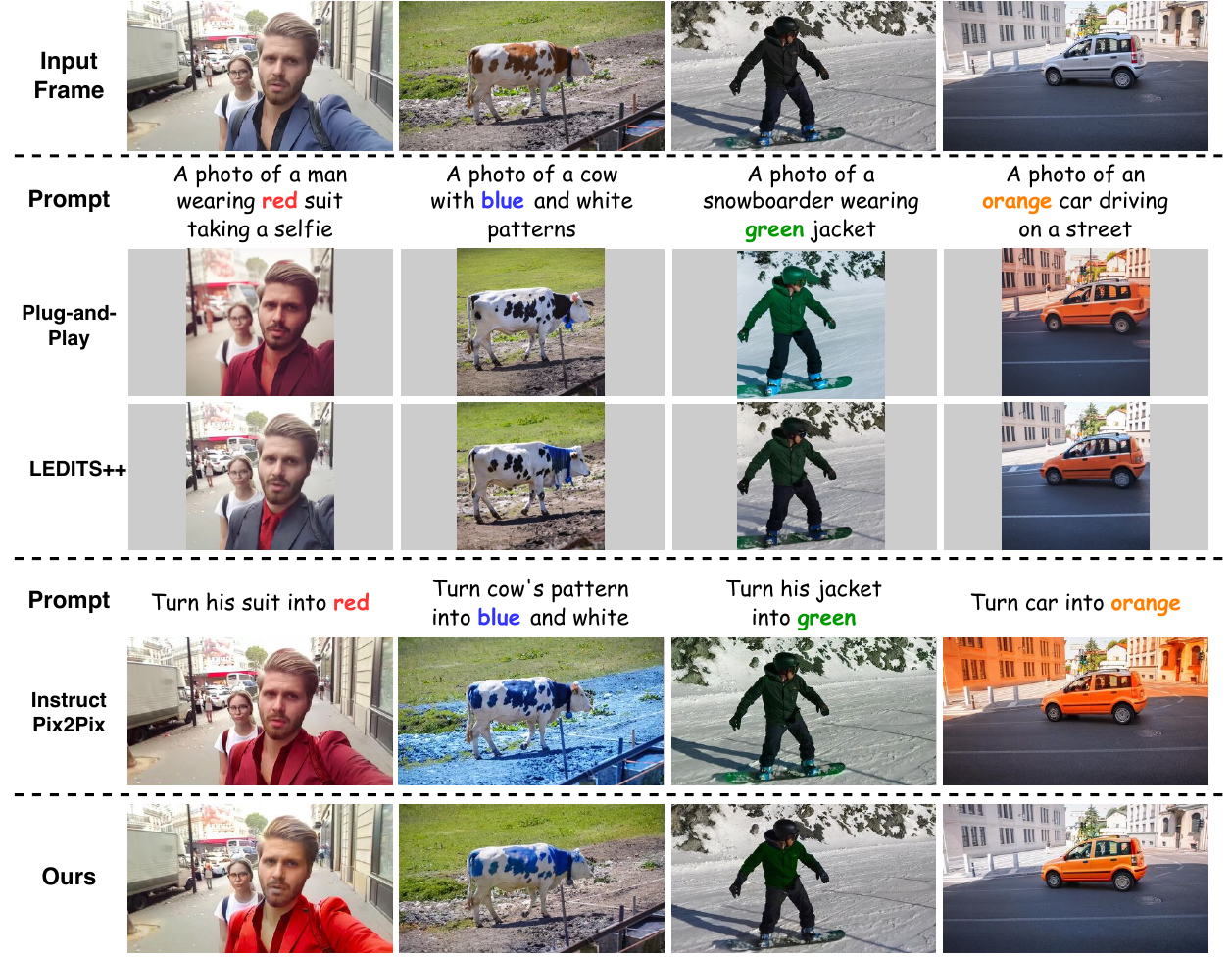}
\vspace{-6mm}
\caption{Qualitative results for intra-frame colour editing: Qualitative comparison of intra-frame colour editing stage between Plug-and-Play \cite{tumanyan2022plugandplay}, LEDITS++ \cite{brack2024ledits}, InstructPix2Pix \cite{brooks2023instructpix2pix}, and our proposed method on DAVIS \cite{ponttuset2017davis} dataset.}
\label{fig:img}
\vspace{-4mm}
\end{figure}

\paragraph{Qualitative Evaluation.}
We evaluate our method on two key stages: intra-frame and inter-frame colour editing, assessing the performance of each step in our video colour editing pipeline, as depicted in Fig. \ref{fig:img} and Fig. \ref{fig:vid} respectively. The results show that our method consistently outperforms state-of-the-art (SOTA) approaches. In the intra-frame stage (see Fig. \ref{fig:img}), our approach achieves superior frame-wise colour editing compared to novel T2I-based image editing models, avoiding common artefacts such as colour bleeding. These artefacts often stem from text prompt ambiguity, including challenges in precisely defining target regions or specific shades of colour, which can result in unfaithful edits. In the inter-frame stage (see Fig. \ref{fig:vid}), T2I-based video editing models, such as FateZero and VidToMe, adapt their frameworks by replacing the spatial self-attention mechanism in T2I diffusion models with temporal-aware cross-frame attention to process temporal information. However, they continue to struggle with flickering artefacts due to limited temporal consistency. While AnyV2V, integrated with InstructPix2Pix, achieves improved temporal coherence by leveraging generative priors from video diffusion models, its results remain compromised by inaccuracies originating from the initial frame editing stage. Our method effectively addresses these challenges, delivering smoother, more consistent edits and showcasing significant advantages in both image and video colour editing tasks.

\section{Ablation Study on Initial Latent Index}
\label{sec:abla2}

\begin{figure}[hbt]
\centering
\includegraphics[width=.95\linewidth]{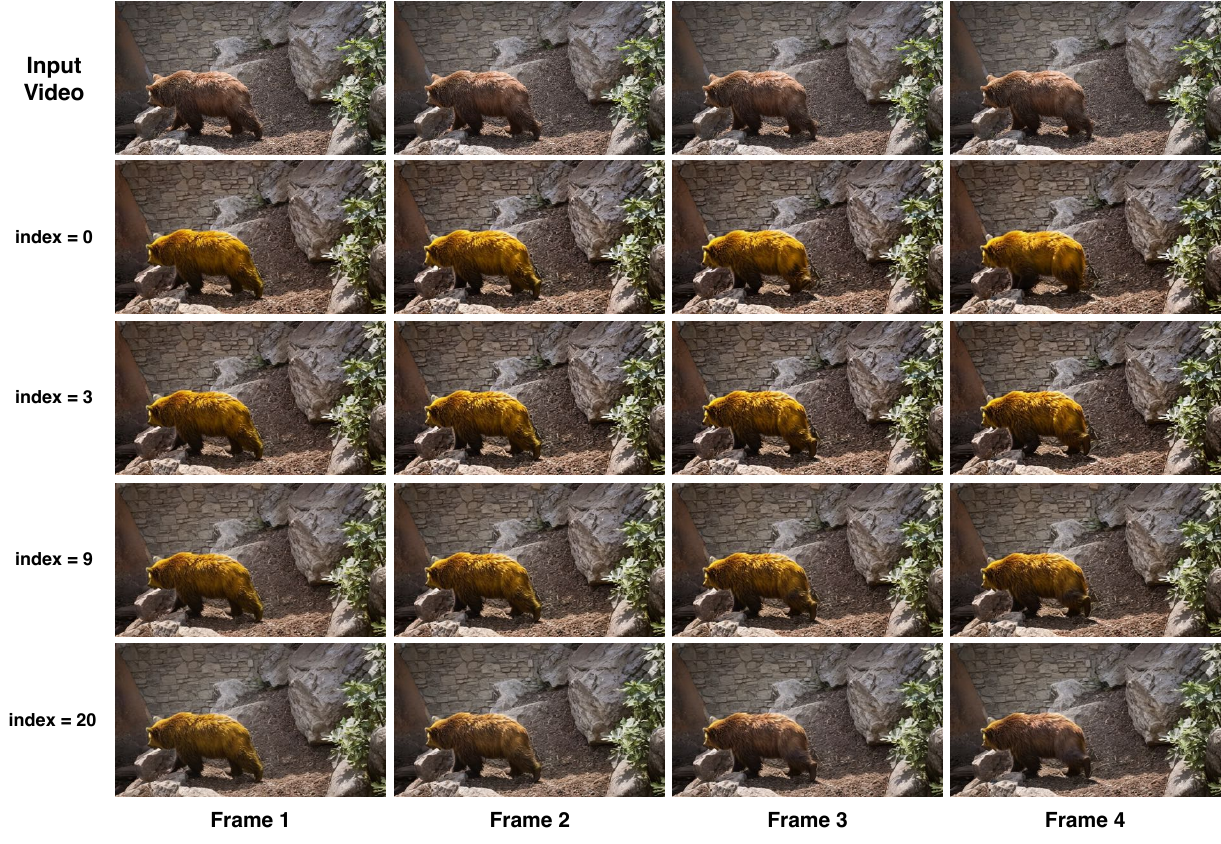}
\caption{Ablation study on the impact of using different initial latent indices (i.e., 0, 3, 9, and 20) on the final colour-edited video.}
\label{fig:s_abla2}
\vspace{-4mm}
\end{figure}

To evaluate the impact of the initial latent index (\(\tau_{idx}\)) on video colour editing, we conducted an ablation study, as presented in Fig. \ref{fig:s_abla2}. This parameter determines the starting point of the sampling process during DDIM inversion, directly influencing the trade-off between semantic detail preservation and colour propagation. Specifically, \(\tau_{idx}\) controls the extent to which the diffusion process relies on the latent information from the initial frame versus the subsequent frames, thereby affecting the consistency of colour edits across the video. In our study, we tested four representative values of \(\tau_{idx}\): 0, 3, 9, and 20. As shown in Fig. \ref{fig:s_abla2}, setting \(\tau_{idx} = 0\) overly relies on the initial latent noise, which degrades the texture quality of the bear in the example video of a bear walking on a rock. This results in a noticeable loss of semantic detail. Conversely, \(\tau_{idx} = 20\) effectively preserves semantic details and texture fidelity but struggles to propagate the colour edits from the initial frame, leading to inconsistent appearance across the video. Intermediate values, such as \(\tau_{idx} = 3\) or \(\tau_{idx} = 9\), achieve a better balance, ensuring semantic details are retained while also maintaining coherent colour propagation throughout the video.

\section{Clarification on Contributions}
\label{sec:contribution}
Our proposed method is not merely a combination of independent modules but a carefully designed framework that strategically adapts specialised approaches to address the challenging task of video colour editing. By redirecting well-established methods trained or tailored for specific tasks, we effectively repurpose them for this domain, achieving a balance between innovation and practicality. Key to our training-free approach is the utilisation of pretrained modules, such as SAM2 \cite{ravi2024sam2} and BLIP-2 \cite{li2023blip2}, which provide robust capabilities for segmentation and multimodal understanding, respectively, enabling precise guidance and interaction in our pipeline.

Furthermore, we tackle complex challenges with efficient and reliable traditional computer vision techniques, such as SLIC \cite{SLIC}, which we adapt to preserve the semantic integrity of the original frame’s background. This adaptation ensures accuracy and efficiency while maintaining computational simplicity compared to more resource-intensive methods. Our design choices are thoughtfully guided by principles aimed at improving the video colour editing process. This demonstrates that, even without training or fine-tuning, our framework can leverage pre-trained modules and diffusion model priors to achieve high-quality video colour editing. Moreover, the training-free nature of our approach not only enables zero-shot capability but also ensures compatibility and scalability, allowing seamless integration with future foundation models and state-of-the-art techniques.


\end{document}